%% file: main.tex
\documentclass[sigconf]{acmart}

\copyrightyear{2019}
\acmYear{2019}
\setcopyright{acmcopyright}
\acmConference[KDD '19]{The 25th ACM SIGKDD Conference on Knowledge Discovery and Data Mining}{August 4--8, 2019}{Anchorage, AK, USA} \acmBooktitle{The 25th ACM SIGKDD Conference on Knowledge Discovery and Data Mining (KDD '19), August 4--8, 2019, Anchorage, AK, USA}
\acmPrice{15.00}
\acmDOI{10.1145/3292500.3330750} 
\acmISBN{978-1-4503-6201-6/19/08}

\usepackage{balance}

\usepackage[utf8]{inputenc}
\usepackage[inline]{enumitem}
\usepackage{subcaption}
\usepackage{array}
\newcolumntype{H}{>{\setbox0=\hbox\bgroup}c<{\egroup}@{}}
\usepackage{mwe}

\def\BibTeX{{\rm B\kern-.05em{\sc i\kern-.025em b}\kern-.08emT\kern-.1667em\lower.7ex\hbox{E}\kern-.125emX}}

\usepackage[font=small,labelfont=bf,textfont=md]{caption}

\begin{document}

\title[Characterizing and Forecasting User Engagement with In-App Action Graphs]{Characterizing and Forecasting User Engagement with In-app Action Graph: A Case Study of Snapchat}

\author{Yozen Liu}
\affiliation{%
  \institution{University of Southern California}
  \city{Los Angeles}
  \state{California}
}
\email{yozenliu@usc.edu}

\author{Xiaolin Shi}
\affiliation{%
  \institution{Snap Inc.}
  \city{Los Angeles}
  \state{California}
}
\email{xiaolin@snap.com}

\author{Lucas Pierce}
\affiliation{%
  \institution{Snap Inc.}
  \city{Los Angeles}
  \state{California}
}
\email{lpierce@snap.com}

\author{Xiang Ren}
\affiliation{%
  \institution{University of Southern California}
  \city{Los Angeles}
  \state{California}
}
\email{xiangren@usc.edu}

\begin{abstract}
While mobile social apps have become increasingly important in people's daily life, we have limited understanding on what motivates users to engage with these apps. In this paper, we answer the question whether users' in-app activity patterns help inform their future app engagement (e.g., active days in a future time window)? Previous studies on predicting user app engagement mainly focus on various macroscopic features (e.g., time-series of activity frequency), while ignoring fine-grained inter-dependencies between different in-app actions at the microscopic level. Here we propose to formalize individual user's in-app action transition patterns as a temporally evolving \textit{action graph}, and analyze its characteristics in terms of informing future user engagement. 
Our analysis suggested that action graphs are able to characterize user behavior patterns and inform future engagement. We derive a number of high-order graph features to capture in-app usage patterns and construct \textit{interpretable models} for predicting trends of engagement changes and active rates. To further enhance predictive power, we design an \textit{end-to-end, multi-channel neural model} to encode temporal action graphs, activity sequences, and other macroscopic features. Experiments on predicting user engagement for 150k Snapchat new users over a 28-day period demonstrate the effectiveness of the proposed models. The prediction framework is \textit{deployed at Snapchat} to deliver real world business insights. Our proposed framework is also general and can be applied to other social app platforms\footnote{\small Code is published at \url{https://github.com/INK-USC/temporal-gcn-lstm}}. 
\end{abstract}





\maketitle

\input{1-intro}

\input{2-action-graph.tex}

\input{3-analysis}
\input{4-model-prediction}
\input{5-exp}
\input{6-related-work}
\input{7-conclusion}



\end{document}

%% file: 1-intro.tex
\section{Introduction}

\begin{figure}[t]
\vspace{-0.0cm}
    \centering
    \includegraphics[width=0.95\linewidth]{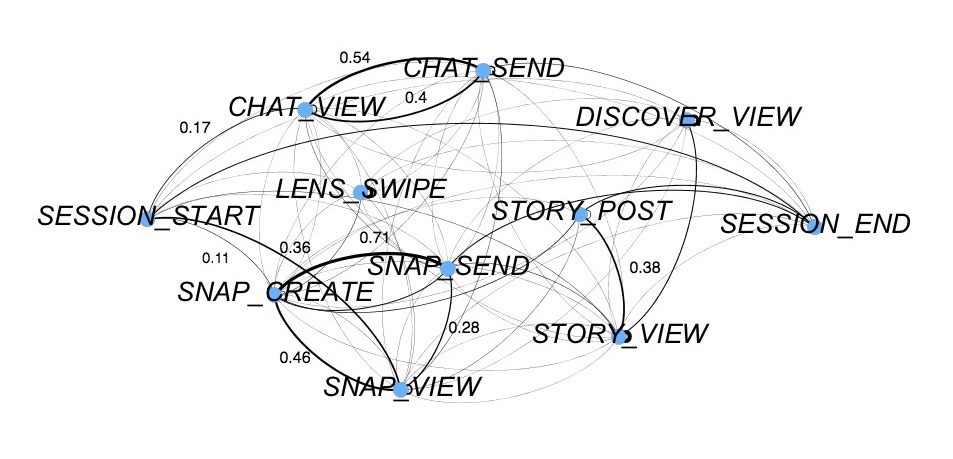}
    \label{fig:dense_graph}
    %
     \vspace{-0.6cm}
    \caption{\textbf{An example \textit{action graph} derived from the in-app activity data of an individual Snapchat user.} Nodes represent actions the user performs when using the app, and edges indicate transition probability from one action to the other within a session (i.e., from open to close the app).}
     \vspace{-0.3cm}
\end{figure}

In recent years there has been a surge of interests on analyzing how users adopt online applications (Apps) on internet and mobile platforms \cite{benevenuto2009characterizing} and understanding what retain user engagement on the Apps \cite{ciampaglia2015moodbar}. Despite many of such studies focus on profiling users \cite{yang2018know} and building more sophisticated models towards surfacing more personalized content for users \cite{lo2016understanding}, however we still have limited understanding on what are the key factors that contribute to user engagement with the Apps---a question that has significant value for retaining and activating users. Therefore, it has become increasingly important to understand and characterize user behaviors and relate them to user engagement (e.g., prediction future engagement trends).

Prior efforts on analyzing user engagement focus on deriving \textit{macroscopic} features from user profile (e.g., demographic information, session statistics)\cite{althoff2015donor} and activity statistics (e.g., time-series of activity frequency) \cite{yang2018know, lin2018ll}.
A major line of existing work identify important characteristics of user behaviors via predicting if a user will return to the platform in a variety of ways. Return rate prediction is one of the most prevalent and significant task on social media platforms as companies are eagerly trying to find out the secret to their success. A few studies aim to predict churn rate, i.e cease activity on the platform, or user retention\cite{au2003novel, kawale2009churn}. Others come in from a different angle to predict when a user will return \cite{kapoor2014hazard}, if a user will return \cite{lin2018ll} or determine the lifespan of a new user \cite{yang2010activity}.
Here we argue that such macroscopic features on user behaviors are (1) insufficient in \textit{characterizing how how users interact with different functions within the Apps} (i.e., fine-grained App usage pattern); and (2) \textit{limited in terms of interpreting the user behaviors} (i.e., lack of explanation by black-box models).

In this paper, going beyond prior work, we propose a new angle to model user behavior with the concept of \textit{Action Graph}---a weighted, directed graph for capturing individual user's in-app action transition patterns (e.g., how likely a user will view message right after she opens the app, and how likely she will create a new message after viewing some).
Action graphs provide microscopic description of user's in-app usage behaviors and thus may contain additional signals to inform future user engagement.
 we take the anonymous data from Snapchat to study user behavior modeling and the connection between action graphs and user engagement. To motivate our study on capturing higher-order graph information, we perform a series of analysis. Through analyzing derivation of lower order features such as graph genes to higher order graph features like user paths, we find the intuition to model complete action graphs as a measure to capture higher-order graph information.

\begin{figure}[t]
\vspace{-0.4cm}
    \centering
    \includegraphics[width=0.83\linewidth]{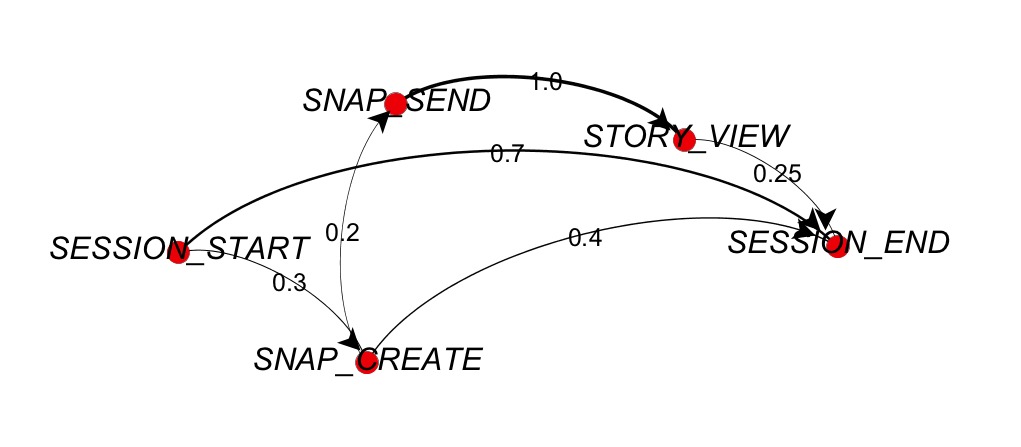}
    \label{fig:sparse graph}
    \vspace{-0.7cm}
    \caption{\textbf{Another example of action graph with fewer number of actions and sparser action transitions (versus the graph in Fig.~1).} }
    \vspace{-0.4cm}
\end{figure}

To enable interpretable user engagement prediction, we first propose a feature based model. Basic features such as macroscopic user level features as well as explainable graph features were modeled in a 2-step fashion. This feature-based model serves as a baseline for our deep neural network model. It also provides valuable insights and explanation to a deeper graph network.

To further improve user engagement prediction, we propose the following models. Activity sequence model with LSTM from previous work as baseline \cite{yang2018know}, static graph model with Graph Convolutional Network, and Temporal action graph model with GCN-LSTM. Temporal graph model best captures time dependencies of action graphs, and outperforms other models and previous baseline. To further improve performance, we propose a deep multi-channel end-to-end model combing activity sequence model, temporal action graph model and macroscopic features. Jointly training all models in an end-to-end manner enables better learning of combined effects and reaches best performance.


\smallskip
\noindent
\textbf{Contributions.}
In summary, we have made the following contributions:
(1) We propose a new \textbf{data model}, \textit{action graph}, for characterizing user in-app behaviors and apply it for forecasting user engagement.
(2) We conduct \textbf{analyses} to understand how different user activity-related signals can help inform their future engagement and find plausible evidences from modeling high-order information in action graphs.
(3) We propose a \textbf{GCN-LSTM model} for learning from temporal action graph and develop a \textbf{multi-channel end-to-end forecasting framework} for integrating with other useful signals.
(4) Extensive experiments on both static and temporal action graphs show that modeling temporal action graphs provide notable performance improvement in user engagement prediction. Our ablation study demonstrate the effectiveness of simple interpretable graph features for predicting user engagement.


\begin{figure}
    \includegraphics[width=0.92\linewidth]{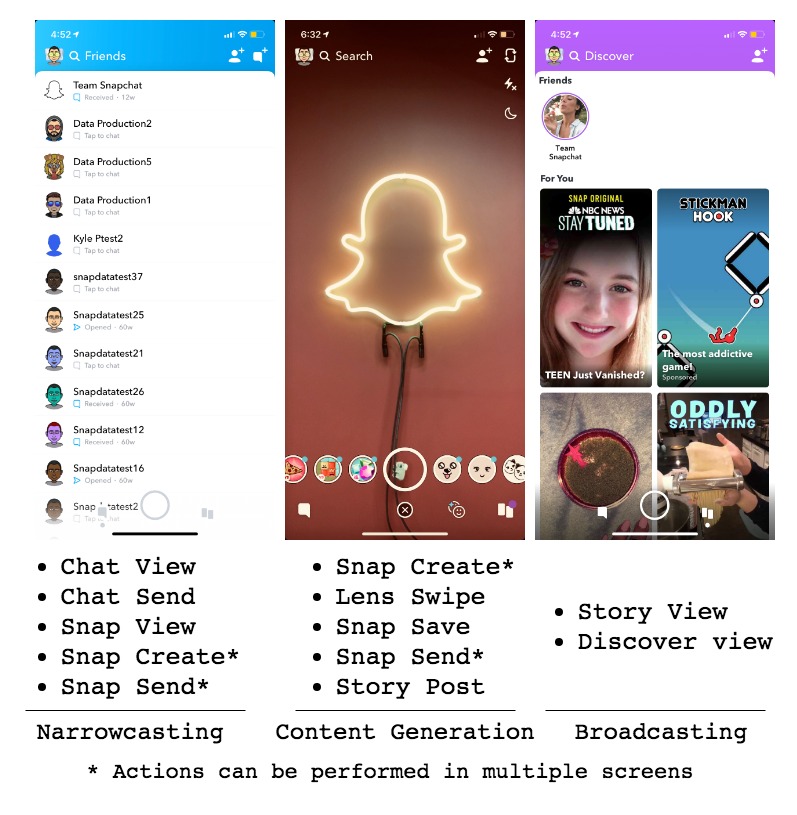}
    \vspace{-0.4cm}
    \caption{\textbf{In-app screenshot of Snapchat and 10 actions to perform as the major functions of the app.} Under each screenshot, we list possible actions users can perform when using the app.}
    \label{fig:screenshots}
    \vspace{-0.0cm}
\end{figure}

%% file: 2-action-graph.tex
\section{Action Graph for Characterizing User Engagement}
In this section we introduce the snapchat dataset used in our study, define the concept of action graph, and provide details on construction of action graphs from user activity data.

\smallskip
\noindent
\textbf{Data Collection.}
We collect the anonymous high level behavioral data of users that registered on the app between April 1st 2018 and September 30th in a specific region. 
A sample of the dataset from over 25 million new users in a specific country over the time period is further extracted.
Four weeks of activity data right after registration is collected for each user, which consists of 10 core in-app functions (shown in Figure \ref{fig:screenshots}) that we find most effective in characterizing user behavior on Snapchat. We use the first two weeks of data for analysis and construction of user action graphs. 
The subsequent two weeks data is used for prediction evaluation of users' future engagement.

\begin{figure}[h]
    \vspace{-0.2cm}
    \centering
    \begin{subfigure}{0.45\linewidth}
        \centering
        \includegraphics[width=\linewidth]{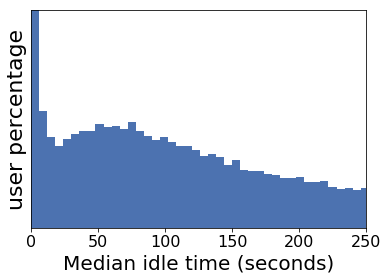}
        \caption{Median idle time distribution}
        \label{fig:idle time dist}
    \end{subfigure}\hspace{0.15cm}
    \begin{subfigure}{0.435\linewidth}
        \centering
        \includegraphics[width=\linewidth]{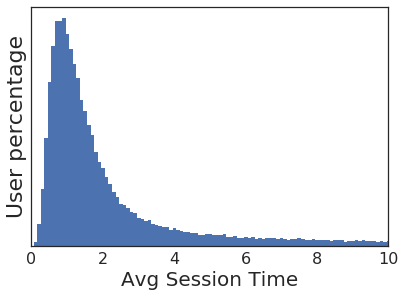}
        \caption{Average session time}
        \label{fig:avg session time}
    \end{subfigure}
    \vspace{-11pt}
    \caption{\textbf{Distribution of user idle time between sessions and average time within sessions.} Note that x-axis in (b) is re-scaled and both y-axes are masked to not show absolute value.}
    \vspace{-0.3cm}
\end{figure}

\smallskip
\noindent
\textbf{Sessions for In-app Activities.}
Ideally, an in-app session is a sequence of continuous actions separated clearly by users' disengagement. However, in real life mobile usage situations signals of starting engagement (i.e. App Open) can be triggered by both opening the app and returning from background mode. Similarly, signal of disengagement (i.e. App Close) can be triggered by disengaging the app or simply enters app-selection mode, checking notification, temporarily back-grounding the app, or replying to messages on another app but return shortly. As we can see, disengagement can be ambiguous as in above situations user will return to engagement from where they've previously left off and resume in a brief period of time. Splitting sessions solely based on disengagement signals recorded such as App Open and App Close will result in ineffectiveness of capturing accurate user action flow and user intention of using the app. Therefore, we need a more accurate definition for in-app sessions. As opposed to an entirely heuristic approach in splitting sessions, previous work has been done to identify sessions by fitting inter-activity time into multiple distributions \cite{halfaker2015user}. By looking at the median idle time per user between start and end engagement signals in Figure \ref{fig:idle time dist}, we can observe that the distribution is a mixture of a long-tail and normal distributions. Median idle time was chosen for the reason that average idle time will subject to extreme outliers. The last 10 percentile of idle time distribution falls under the 25 seconds mark, which is coincidentally where two distributions intersect. We accordingly define 25 seconds as our threshold idle time to split sessions in addition to the disengagement signals (i.e. App Open/Close) in our experiments. Figure \ref{fig:avg session time} shows the distribution of the time spent within each session on Snapchat following our definition of in-app sessions.

\smallskip
\noindent
\textbf{Definition of Action Graphs.}
In this study we define a new concept \textit{Action Graph} to model user behavior. An action graph is a directed graph with 10 in-app actions (shown in Figure \ref{fig:screenshots}) plus Session Start and Session End as nodes. Similar to a Markov chain model, its edges are transition probabilities between actions nodes. Start and end nodes, i.e. Session Start and Session End, have only out-edges and in-edges respectively. Each user has his/her own unique action graph in the observed period. In the case of modeling temporal graphs, a unique action graph is derived at each time step per user. Examples of action graphs are shown in above figures, where Figure 1 serves as an example of a more engaged user and Figure 2 a less engaged user.

\smallskip\noindent
\textbf{Action Graph Construction.}
In order to construct a meaningful action graph for each user, We first split a stream of core actions listed in Figure \ref{fig:screenshots} into partitions of sessions that best captures the user's behavior on Snapchat. Each user has one or more session sequences. We remove sessions containing no actions but \textit{Session Start/End} (i.e. invalid sessions). In order to make transition probability of action graphs more meaningful for each user, we only keep users with 5 or more valid sessions in our analysis, which results in around 150K sampled users for our study. Each user averages around 7 sessions a day and 98 sessions in the first two weeks of the observation period. 

Specifically, each action graph is build with \textit{all valid sessions of a user} using her/his activity data. 
Each \textit{edge} in an action graph is calculated as the transitional probability between two actions aggregated in all sessions of a user. Action graphs can have a maximum of 12 nodes including Session Start and Session End. 



%% file: 3-analysis.tex
\section{Large-scale Data Analysis}

\subsection{Action Type}
As we show in Figure \ref{fig:screenshots}, users' interactions with core functions on Snapchat are in three major categories: content generation, narrowcasting and broadcasting. Narrowcasting actions are triggered by one-to-one communication to interact directly with users and their friends, including Snap Send, Snap View, Chat Send and Chat View. Broadcasting behaviors are those that consumes and shares content to the entirety of a users' network. Such actions include story related activities and discover story activities such as Story Send, Story View and Discover View. The user interface of Snapchat is designed in a way that most narrowcasting functions are located on the left side of the start screen while most broadcasting functions on the right side. 

The purpose of most social media studies is to understand user engagement, content consumption and to increase or retain the level of engagement and consumption. Preferably, a more engaged user will more likely to have more transitions between narrowcast and broadcast, consumption and creation activities. A transition from narrowcast actions to broadcast actions is  favorable because consumption of broadcast contents bring in ad and revenue while users' use the app to communicate with friends. Observing transitions from broadcast to narrowcast can indicate that a user is more engaged with their friends and more likely to be retained. We find that if a session starts with narrowcasting activities, one third leads to broadcasting activities. If a session starts with broadcasting activities, one fifth of the time it leads to narrowcasting activities. It is fairly reasonable that a user only completes their intended communication or consumption activities and doesn't imply that users are not actively engaged. The subject matter here simply is how to increase engagement for those less engaged by encouraging them to navigate through cross functionality.

\begin{figure}[h]
    \centering
    \includegraphics[width=0.5\linewidth]{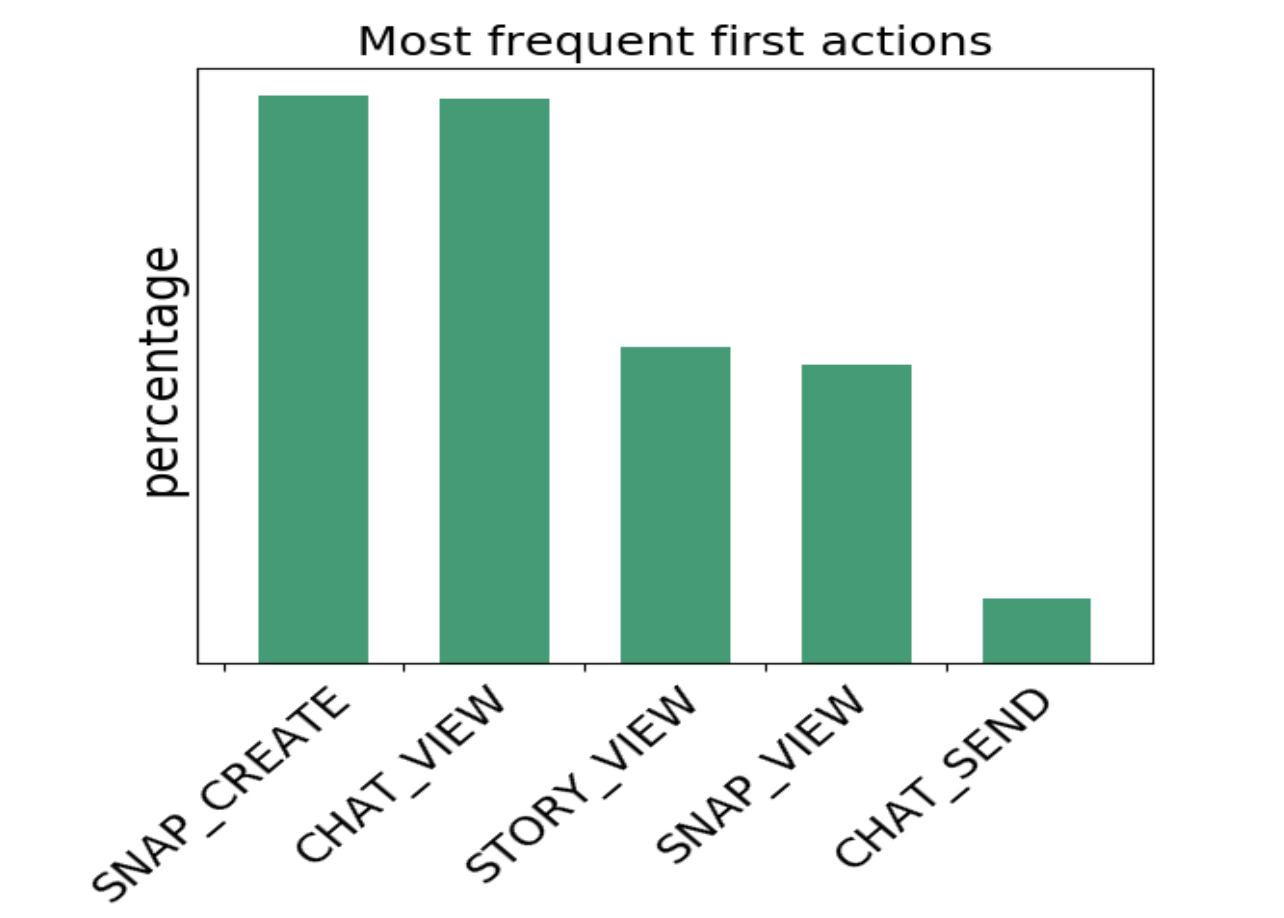}
    \vspace{-10pt}
    \caption{\textbf{Most frequent first actions after Session Start.} Figure shows snap create and chat view are two actions that most frequently triggers engagement. Y-axis is masked in order not to show absolute values.}
    \label{fig:graph first}
    \vspace{-10pt}
\end{figure}

In order to better understand what motivates users to start engaging with Snapchat, we examine the first action after \textit{Session Start} in each session. Interestingly, we can see from Figure \ref{fig:graph first} that the most frequent first action for users are Snap Create (a content generation action) and Chat View (a narrowcasting action). Following by Snap View (which is another narrowcasting action) and Story View (a broadcasting consumption action). Additionally to actions that trigger users to engage, we also look into the duration of time a user spend on the app each session. In Figure \ref{fig:avg session time} we find that the average session time per user shows a normal distribution and peaks within several minutes.


\subsection{Session Gene}
\begin{figure}[h]
    \centering
    \includegraphics[width=.75\linewidth]{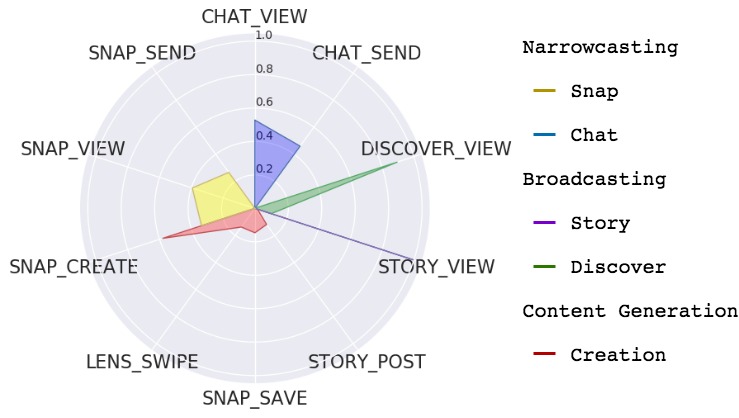}
    \vspace{-5pt}
    \caption{\textbf{Session genes.} We derive 5 session genes and chart shows the probability composition of each gene.}
    \label{fig:topic radar}
    \vspace{-0.3cm}
\end{figure}

\begin{figure}[h]
\centering
\begin{subfigure}[]{.5\linewidth}
  \centering
  \includegraphics[width=.85\linewidth]{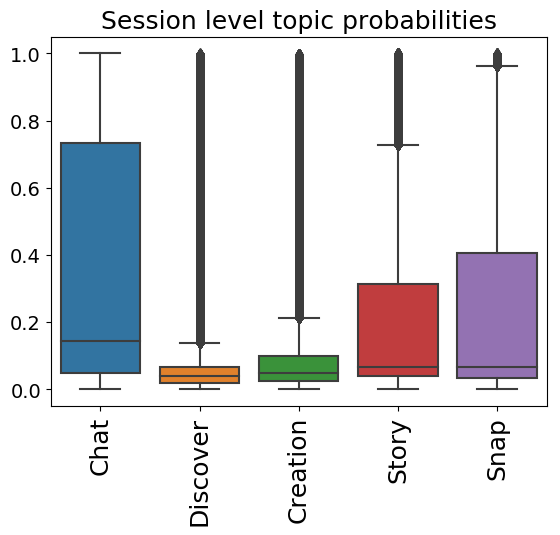} 
  \caption{Session level}
  \label{subfig:session topic}
\end{subfigure}\hspace{0\textwidth}%
\begin{subfigure}[]{.5\linewidth}
  \centering
  \includegraphics[width=.9\linewidth]{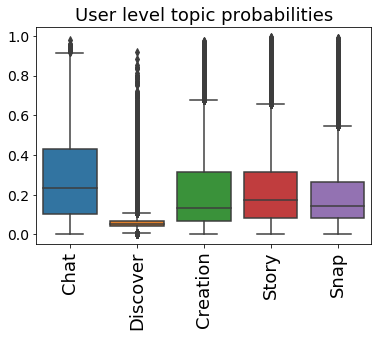} 
  \caption{User level}
  \label{subfig:user session topic}
\end{subfigure}
\vspace{-10pt}
\caption{\textbf{Session gene distributions on session and user level.} Aggregation on different level is used to compare and provide analysis insight.}
\label{fig:topic}
\vspace{-0.1cm}
\end{figure}

In order to have a deeper understanding of the sessions of action sequences and better connecting users intent of using the app (i.e. content generation, narrowcasting and broadcasting), we break sessions into session genes by applying soft clustering method Latent Dirichlet Allocation (LDA) \cite{Blei2003LDA}. Similar to topic modeling, we take each action as a word, each session as a document, and we name the topics learned from LDA  ``session genes''. Session genes are fundamental elements that characterize the functions of a session at a finer level. To apply LDA to our data, we consider each session sequence as a single document and each action as a word. Five topics were derived from the LDA model and the topics are interpretable by summarizing the words with top weights in a topic. Each session contains a mixture of topics therefore we can view them as ``session genes''. The 5 genes derived are: Chat, Snap, Story View, Discover View and Content Creation. Figure \ref{fig:topic radar} shows which actions comprise each gene. The content creation gene contains the widest span of actions, these actions can imply that users are sometimes playing around with the app and creating content while not necessarily sending them. Session genes can perfectly characterize the composition of each session that forms our action graphs. From Figure \ref{subfig:session topic} we can observe that genes of communication eg. `Chat' and `Snap' are the most probable and common amid all 5 genes. We can conclude that narrowcasting communication features are the most used following with broadcasting (Story View). Next, we aggregate on a user level with all sessions of a user then take the mean of all users. In Figure \ref{subfig:user session topic} topic probabilities appears further normalized and this indicates that the majority of Snapchat users have a balanced usage across all functions in the app. Comparing session level aggregation Figure \ref{subfig:session topic} and user level aggregation \ref{subfig:user session topic}, we can see that users that uses narrowcasting communication functions (Chat and Snap) more on Snapchat has a higher session count on average.


\smallskip
\noindent\textbf{Common paths.}
Common paths are also the fundamentals of building action graphs, as the transition probabilities of graph edges are calculated from these paths. Common paths are derived from session sequences by extracting bi-grams, tri-grams and so on.
Conversely to session genes which each session sequence are viewed as a bag of words in LDA model fit, common paths preserves the order of action sequences. As the fundamental building block of action graphs, common paths help us interpret action graphs and how it captures user behavior pattern. 

\smallskip

\subsection{User Clusters and Engagement}
In this section, we are trying to answer a simple question: How do the behavioral patterns described above correlate with user engagement? Enable to demonstrate the necessity to model higher order action graphs, we start from correlating lower level features to user engagement. We attempt user clustering with feature sets at each level. The number of clusters is determined by silhouette analysis. We pick the largest silhouette score and set cluster number to 4. Number of clusters are kept consistent to enforce comparability.

\begin{figure}[h]
    \centering
    \includegraphics[width=0.285\linewidth]{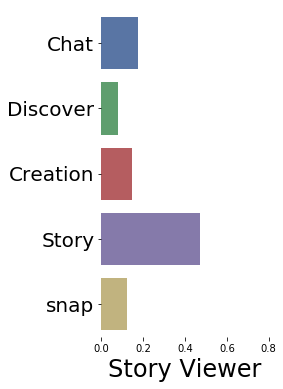}
    \includegraphics[width=0.2\linewidth]{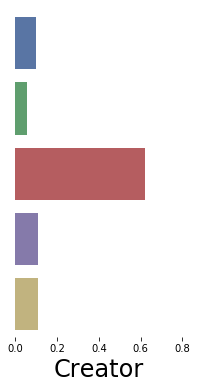}
    \includegraphics[width=0.2\linewidth]{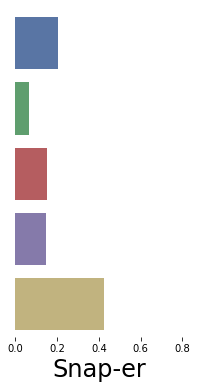}
    \includegraphics[width=0.2\linewidth]{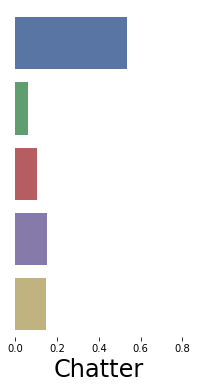}
    \vspace{-5pt}
    \caption{\textbf{Session gene profiles}. We derive 4 user clusters based on user's session gene constitution (i.e., probability of holding each session gene), and name them as: Story viewer, creator, snap-er, and chatter clusters.}
    \label{fig:gene cluster}
    \vspace{-0.1cm}
\end{figure}

\smallskip
\noindent\textbf{Session gene clustering.} 
To cluster lower level graph features, we apply K-means clustering on probabilities of session genes. With 4 clusters we are able to separate users into groups with specific dominating genes. As a result groups of users with dominating story, creation, snap and chat genes can be separated as shown in Figure \ref{fig:gene cluster}. However, As we can see in Figure \ref{subfig:Session gene cluster active rate}, these user clusters are able to separate users of different engagement level but a clear difference cannot be seen.

\smallskip
\noindent\textbf{Common paths clustering.} 
With higher order features such as common paths, we repeat the same method and steps to cluster users. Using the probability normalized by session count and length per user, we derive 4 user clusters. Each cluster has a different distribution over the top paths with groups of users favoring particular paths . Nonetheless, the separation of engagement rate in Figure \ref{subfig:top pattern cluster active rate} for each user cluster are still not clear. Therefore, we move on to higher level features.

\begin{figure}[h]
    \vspace{-0.2cm}
    \centering
    \includegraphics[width=0.5\linewidth]{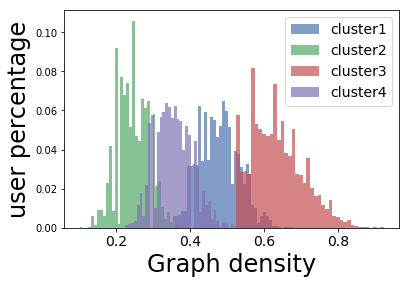}
    \vspace{-10pt}
    \caption{Density distribution of clusters by graph metrics}
    \label{fig:graph cluster density}
\vspace{-10pt}
\end{figure}

\smallskip
\noindent\textbf{Graph metrics clustering.} 
Once Again, we repeat previous clustering steps but replace with action graph level metrics, i.e. number of nodes, number of edges, and graph density, as features. Figure \ref{fig:graph cluster density} shows the graph density distribution of users clustered by their graph metrics. Different from lower level feature user clusters, we see a clear separation between user engagement rate in Figure \ref{subfig:graph cluster active rate}. Obviously, different user types are most efficiently captured with the use of action graphs and its simple features.

In summary, action graphs contain all the above information and is far more informative regarding user engagement. Therefore motivates the use of action graph modeling to predict future engagement.

%% file: 4-model-prediction.tex
\section{Forecasting User Engagement}
In this section, we define two prediction tasks related to forecasting of user engagement, and propose a series of forecasting models for solving the two tasks, based on both \textit{hand-built, explainable features} and \textit{deep neural networks}. We start with a feature-based model that leverages features of different granularity, and move on to a more complex end-to-end prediction model to demonstrate action graphs as a good predictor of future user engagement level.

\subsection{Task Formulation}

We formulate a novel prediction task to predict the future engagement of a user. The prediction task leverages previous discovered insights and demonstrates that action graphs contributes to more accurate predictions. The prediction task can be formulated in two ways, as a classification problem, or as a regression problem.

\smallskip
\noindent \textbf{1. Classification of user engagement trend.}
First we define a 3-class classification task. In the following 2 weeks after observation period after registration, ie. week 3 - week 4, we record the days that a user is active. An active day is defined as a user having at least one valid session, which is a session containing at least one action we record. If a user has more active days in week 3 - week 4, we define that its engagement trend ‘Increases’ and vice versa ‘Decreases’. If active day counts are the same from the first two weeks to the following two weeks, the user will be labeled as another class which stays the same. We also observe that in week 3 - week 4, active days appears to be more polarized than the first two weeks.

\smallskip
\noindent \textbf{2. Prediction of user active rate.}
We formulate active rate prediction as a regression problem. Active rate is defined as “\# of active days / \# of total days” in the coming weeks. This task aims to predict the active rate of the coming weeks rather than only predicting the change of engagement trend between the two periods.

\begin{figure}[t]
\vspace{-0.1cm}
\centering
\begin{subfigure}[]{.32\linewidth}
  \centering
  \includegraphics[width=\linewidth]{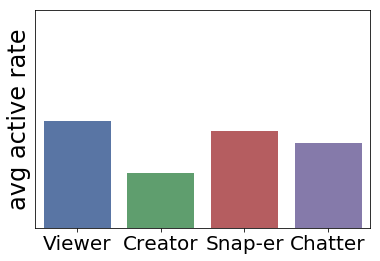} 
  \caption{Session gene}
  \label{subfig:Session gene cluster active rate}
\end{subfigure}\hspace{0\textwidth}%
\begin{subfigure}[]{.32\linewidth}
  \centering
  \includegraphics[width=\linewidth]{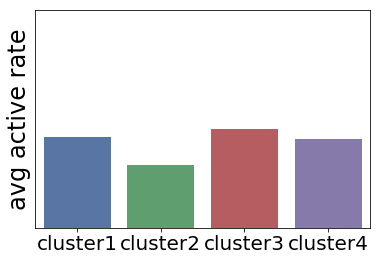} 
  \caption{Common paths}
  \label{subfig:top pattern cluster active rate}
\end{subfigure}
\begin{subfigure}[]{.32\linewidth}
  \centering
  \includegraphics[width=\linewidth]{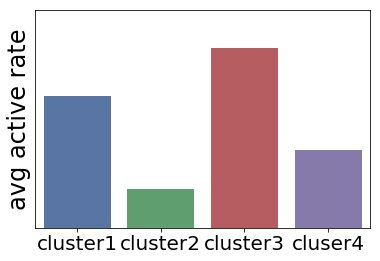} 
  \caption{Graph level}
  \label{subfig:graph cluster active rate}
\end{subfigure}
\vspace{-10pt}
\caption{\textbf{Average active rate of different clustering.} Figures show that clustering with graph level features most correlates to active rate.}
\label{fig:clusters active rate}
\vspace{0.0cm}
\end{figure}

\subsection{Feature-based Forecasting Model}

We first propose a feature based model with macroscopic features and explainable graph features. The feature based model provides interpretability to deep neural graph models.

\smallskip
\noindent \textbf{Macroscopic Features.}
Macroscopic features are extracted from user profiles with obvious features like session count and average session time they are used to provide a baseline prediction. These features include \begin{enumerate*} \item The average number of sessions per day for a user \item The average time user spend on Snapchat per session \item The gender of a user \item The maximum age of a user during time period\item The accumulated friend count of a user \end{enumerate*} Macroscopic features are easy to understand and explain how it could affect the active rate.

\smallskip
\noindent \textbf{Explainable Graph Features.}
An action graph can contain a vast amount of user behavior pattern information that is hard to see from a high level action graph. Therefore, we derive a few explainable graph features from action graphs to show that they are able to capture user behavior information and provide insights to user activity pattern and the way they interact with the app. \begin{enumerate*}
    \item \textbf{Most likely first action}: The first action after session start. First actions often tells us what the user’s most favorable interaction with Snapchat is, or what triggers the user to start using the app.
    \item  \textbf{Most likely action probabilities}: The most likely last actions before session end.
    \item \textbf{K-hop paths}:  A k-hop path starts from session start. K-hop paths are most likely action paths a user can take by calculating the joint probability $\Pi P(T_i)$  of transition probabilities $T_i$ of edges along the path.
    \item \textbf{End-to-end paths}: Here we utilizes BFS search to find paths from start node to end node. A path length cap is set to 6 to prevent indefinite search in loops and self-loops. We calculate the strength of each path (different length possible) as  $P^{1/N}$  where $P$ is the joint probability  $\Pi P(T_i)$  of edges along path, and $N$ the number of edges in path.
    \item \textbf{Strongest cycles}: We utilize Johnson’s algorithm to extract all elementary cycles from a graph.  We then calculate cycle strength as $P^{1/N}$ where  $P$  is the joint probability $\Pi P(T_i)$ of edges and $N$ is the number of edges in cycle.
\end{enumerate*}
We will later show in section \ref{sec:feature selection} selected graph features that are indicative to user engagement.

\smallskip
\noindent \textbf{Predictive Models over Features.}
The feature-based model combines simple features vectors as input to the classifiers. The classifiers that we experimented on are SVM and Softmax classifier. As a regression problem we experimented on Linear regression and Ridge regression methods and select the optimal performance. SVM serves as a baseline where Softmax classifier reaches the best performance when combining all features.


\subsection{Deep Neural Forecasting Models}

To further improve user engagement prediction, we employ deeper neural models for various input signals. We utilize LSTM to model activity sequence time series data as baseline. 
We then propose static and temporal graph modeling methods and a multi-channel end-to-end training framework as our final best model for engagement prediction.

\smallskip
\noindent \textbf{Activity Sequence model.}
Inspired by previous work that models user sequential behavior data and utilize LSTM sequence-to-sequence learning technique to encode and predict churn rate\cite{yang2018know}. We implement 2 layer LSTMs with activity sequence containing actions in Figure \ref{fig:screenshots} as a 10 dimensional time series input. The LSTM structure is able to model time series behavioral data and capturing evolvement of user activities. Each layer of the LSTM computes the following transformations:
\begin{flalign}
    \label{eq:lstm}
    &f_t = \sigma (W_f \cdot [h_t-1, x_t] + b_f) \\
    &i_t = \sigma (W_i \cdot [h_t-1, x_t] + b_i) \nonumber \\ 
    &c_t = f_t * c_t-1 + i_t * tanh (W_c \cdot [h_t-1, x_t] + b_c) \nonumber\\
    &o_t = \sigma(W_o \cdot [h_t-1, x_t] + b_o) \nonumber\\
    &h_t = o_t * tanh(C_t) \nonumber
\end{flalign}
\noindent where $t$ is the time step in terms of days. $h_t$, $c_t$, $x_t$ are the hidden state, cell state, and previous layer hidden state at time $t$. $f_t$, $i_t$, $o_t$ are respectively the  forget gate, input gate and output gate.

In our implementation of LSTM training, we apply softmax function $\text{softmax}(x_i) = \frac{\text{exp}(x_i)}{\sum_{j} \text{exp}(x_j)}$ to the linear projection of LSTM output $o_T$ for our classification prediction $\hat{y}$ (\ref{eq:lstm softmax prediction}). Dropout is also applied to avoid over-fitting.
\begin{flalign}
    \label{eq:lstm softmax prediction}
    \hat{y} = \text{softmax} (W_c o_T + b_c).
\end{flalign}

Since LSTM can successfully capture time dependencies on user behavioral data, we will further combine its output embedding to other features and devise a more robust prediction model.

\smallskip
\noindent \textbf{Static graph model.}
Static graphs are aggregated over the entire 2 weeks observation period, each user has a single unique graph. Although our derived explainable graph features serves well representing the graph, it does not capture all dependencies and pattern. Therefore, we employ a more powerful way to encode static graphs and conduct prediction. Graph Convolutional Network (GCN) \cite{kipf2016semi} encodes each node as an embedding vector, can perform in a semi-supervised or supervised scenario to classify nodes. A GCN updates the node embedding using its neighbor information at each layer with message passing, and learn the representation of node. The layer-wise propagation rule in vector form is defined as
\begin{flalign}
    \label{eq:GCN}
    h_{i}^{l + 1} = \sigma (\sum_{j} w_{ji} h_{j}^{l} W^l).
\end{flalign}
Where $h_{v_i}$ is the feature of node $i$, $j$ are the neighboring nodes of $i$, and $W^{l}$ is the weight matrix of the $l$-th layer, $\sigma$ can be a non-linear activation such as ReLU. 

\begin{figure}[h]
    \vspace{-10pt}
    \centering
    \includegraphics[width=1.01\linewidth]{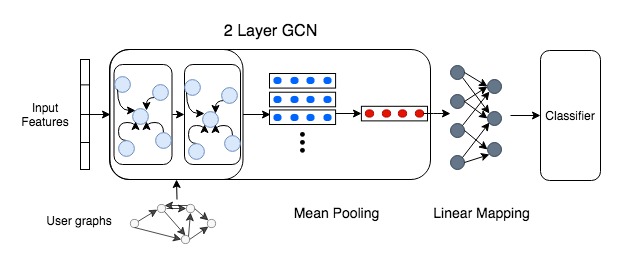}
    \vspace{-0.9cm}
    \caption{\textbf{GCN model over static graph for engagement forecasting.} User's static action graph (along with node features) are inputs to a two-layer GCN structure with mean-pooling after the last layer. Single graph embedding vector is then linearly mapped then feed to classifier.}
    \label{fig:gcn}
    \vspace{-0.1cm}
\end{figure}

We modify a two layer GCN structure to accommodate our case of graph classification on directed graphs as shown in Figure \ref{fig:gcn}. To classify small graphs of many users, we train all small graphs together as a batched large graph. Utilizing the DGL library, graphs can easily be batched together. At each pass a mean-pooling transformation $v_{G} = \frac{1}{n} \sum^{n}_{i} v_{i}$ is applied on the output of two layer GCN to transform all node embeddings of a graph into a single graph representation. This will result in each graph having an unique embedding vector that characterize the structure of the graph. We then apply a softmax layer to linear projection of graph embedding vector $v_G$ for our classification prediction $\hat{y}$. 
\begin{flalign}
    \label{eq:graph softamx predict}
    \hat{y} = \textit{softmax}(W_c v_G + b_c).
\end{flalign}
GCN graph embedding is a suitable representation of action graphs for the reason that it is trained and learned with our target engagement information and also incorporates neighboring information in the action graph at each pass. We can further combine such graph embedding with other features for stronger prediction.

\begin{figure}[b]
    \vspace{-0.2cm}
    \centering
    \includegraphics[width=.82\linewidth]{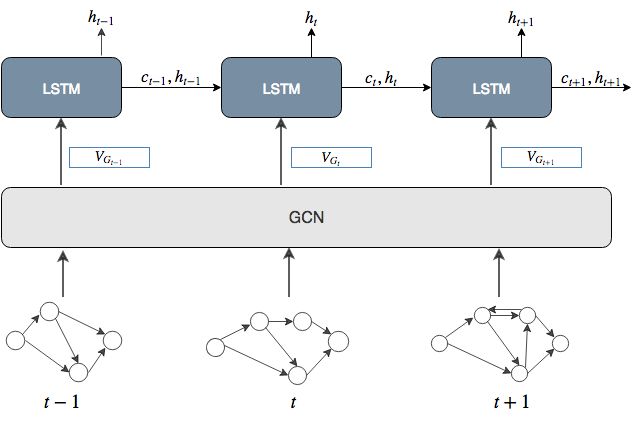}
    \vspace{-0.5cm}
    \caption{\textbf{GCN-LSTM model over temporal action graph (i.e., a series of action graph snapshots) for engagement forecasting.} At each time stamp we input the action graph snapshot to a GCN module and feed its output embedding (after mean pooling) to a LSTM for prediction.}
    \label{fig:temporal}
    \vspace{-0.2cm}
\end{figure}

\smallskip
\noindent \textbf{Temporal Graph Model.}
In real world scenarios, action graphs evolve in a temporal fashion. Previously we model action graphs as static graphs ie. aggregate all sessions in observation period to derive a single graph. Here we introduce a time dependent variant of action graph modeling - temporal graph modeling. 

While static graphs fail to capture the evolution of user behavior, temporal graphs account for the change in user pattern at each time step. We aggregate all sessions between each time step and computes an action graph. In our model, a day is used as the unit of time step. Thus, for a 14 day observation period we can derive 14 separate temporal graphs. A Long short-term memory network [eq \ref{eq:lstm}] fits perfectly in this scenario for the reason that it recognizes temporal dependencies. In our model, each graph is transformed by Graph Convolutional Network(GCN) to a single graph embedding vector. We utilize a single GCN to encode all graphs at all time steps. Graph embedding produced by GCN at all time steps can be viewed as a sequence. This sequence is subsequently input to a single LSTM network for prediction. In simple terms, we compute an action graph at the end of each day with sessions in the day for each user, derive graph embedding of all graphs and feed to LSTM network as a timely sequence. A softmax layer is then applied to the last hidden state output of LSTM for classification task. 

\smallskip
\noindent \textbf{Deep Multi-channel Forecasting Model.}
We propose a more intuitive and desired way to train the model by combining the LSTM for activity sequence and GCN action graph model along with macro features at the last stage to compute a single loss and back propagate to all models. 
Cross-entropy loss for each class $c$ in softmax classifier can be calculated as
\begin{flalign}
    l_c = - \sum_{c} y_{i, c} \text{log} (\hat{y_{i, c}}).
\end{flalign}
for each user $i$ and its binary ground truth $y_{i, c}$ if $y_i$ is class $c$. Alternatively, a linear SVM classifier can be used in place of softmax classifer. The loss will then be calculate as multi-class hinge loss 
\begin{flalign}
    l_i = \frac{1}{N} \sum^{N}_{i\neq y_i} \text{max}(0, 1 - w_{y_i}x + w_{j}x).
\end{flalign}
where $w$ is model parameters. For our regression task we simply apply a linear regression layer and its is computed as Mean Squared error $l = \frac{1}{N} \sum^{N}_{i} (y_i -\hat{y_i})^2$.

Loss is then back propagated to the end-to-end framework into both models to complete the training process. By training in an end-to-end fashion the models learn together and reaches best performance.

\begin{figure}[h]
\includegraphics[scale=0.36]{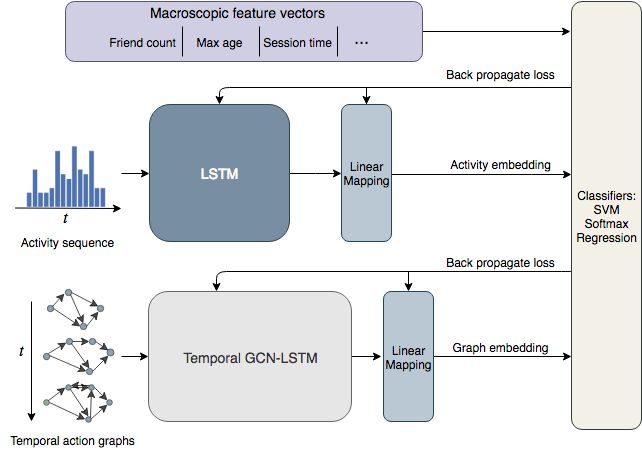}
    \vspace{-0.1cm}
    \caption{\textbf{Deep Multi-channel Forecasting Framework.} In our final forecasting model, macroscopic feature vector is combined with LSTM model (for activity sequences) and the proposed GCN-LSTM model (for temporal action graph). All these modules are jointly trained in an end-to-end manner for user engagement prediction.}
    \label{fig:end-to-end}
    \vspace{-0.1cm}
\end{figure}

Figure \ref{fig:end-to-end} illustrate a combination of macroscopic features, LSTM activity sequence and temporal GCN-LSTM model. We find that jointly training these models end-to-end result in best performance predicting user engagement.

%% file: 5-exp.tex
\section{Experiments on User Engagement Prediction}
We conduct our experiments by adding features one by one to explore the effectiveness of each feature. We include 2 weeks activity sequence data modeling of our full observation period along with 2 weeks user graph data to ensure comparability. We aim to predict two variants of future engagement: \begin{enumerate*}\item User engagement trend \item Future active rate\end{enumerate*}. We conduct experiments with different combinations of features as our model input, and predict a single class or value for the two tasks respectively. Intuitively, we would expect to see that graph embedding performs better than explainable graph features also giving a boost to existing activity sequence modeling method. Combining activity sequence embedding, temporal graph embedding and macroscopic features reaches best performance.



\subsection{Datasets and Experimemt Setup}
All experiments were done on the same sub-sampled data set of new users extracted from the Snapchat database. The original new user data to is sub-sampled into a 1:1:1 split for each of the three classes of engagement trend to avoid a skewed distribution. Labels of engagement trend were determined by the comparison of active days in first 2 weeks and the following two weeks; Labels of active rate for regression task is the active days of the following two weeks to the first two. Evaluation is done by randomly splitting data into 80:20 training and testing set for 10 times, then averaging the 10-fold cross validation score. The experiments were done on a single google cloud engine machine with only CPUs. 

\subsection{Model Training and Hyperparameters}
Our experiments were done with SVM, softmax and linear regression as the final prediction layer in our model.
All variables are also scaled before training using methods robust to outliers. For our GCN network we employ a two-layer configuration, applying mean pooling at the last layer to obtain a single vector representation for each action graph. Adjacency matrix for each graph is used as input feature. We configured our LSTM network as 2 layers with dropout (0.5) and embedding size of 32 which is empirically determined from previous work and further verified in this study. We utilize DGL library \footnote{https://www.dgl.ai/} and PyTorch \footnote{https://pytorch.org/} for implementation of GCN, LSTM models and our end-to-end framework. In our experiments, we did not pre-train the LSTM and GCN models as we found training them end-to-end from scratch yield similar (or slightly better) results.

To determine hyper parameters for our model, we conducted grid search using a held-out validate set in our experiments. We apply $0.5$ dropout to LSTM and a L2 regularization ($1e-3$) to the classifier layer to avoid model overfitting. 
Hidden state size of GCN was set to 20 to avoid overfitting as the action graph is relatively small in our case.
Other input settings to our GCN network such as initializing with random Xavier normal/uniform initialization was also experimented but reaches similar but not better results.

\subsection{Evaluation Metrics}

\noindent \textbf{Metrics for evaluating engagement trend classification.}
For our 3-class classification problem of engagement trend prediction, we use F1 score as our evaluation metric. In addition to accuracy, F1 score takes false positives and false negatives into account, hence a better evaluation metric. F1 score is the harmonic mean of precision and recall. Where precision = $tp/(tp+fp)$ and recall = $tp/(tp+fn)$. F1 score is then computed as $2*(precision*recall)/(precision+recall)$. All reported F1 score is Macro F1 score, which computes the average of per-class F1 metrics in order to deal with inbalanced numbers of instances in each class.

\noindent \textbf{Metrics for evaluating active rate regression.} For our active rate regression problem, we compute Root mean square error RMSE as our evaluation metric. $RMSE = \sqrt{\frac{\Sigma_i(\hat{y_i} - y_i)^2}{n}}$. Comparing to another common metric for regression task, Mean absolute error, RMSE penalizes larger incorrect predictions more.

\begin{table}[h]
\caption{\textbf{Performance comparison on classification of user engagement trends.} We performed 10-fold cross validation to compute the F1 score (based on precision and recall). We applied two different classifiers for feature-based models, end-to-end neural models, and their variants.}
\vspace{-0.2cm}
\label{tab:activity prediction}
\centering
\scalebox{0.95}{
\begin{tabular}{@{}lccH@{}}
\toprule
\textbf{Models}                                       & \multicolumn{2}{c}{\textbf{F1 score}}                               \\\hline
\textit{Feature-based model}                                     & SVM                   &  Softmax              & XGBoost                     \\ \midrule
Macroscopic features (Macro)                                      & .372                        & .417                           &.491                             \\
Graph features                                & .473                       & .473                            &.503                             \\
Macro + Graph features                                    & .479                        & .480                            & .529                            \\ \midrule
\textit{End-to-end neural model}                                 &                         &                             &                             \\ \midrule
static GCN~\cite{kipf2016semi}                                      &.490                         & .516                            &                             \\
Macro + static GCN                            &.497                         &.515                             &                             \\
activity LSTM~\cite{yang2018know}                &.611                         & .614                            &                             \\
Macro + activity LSTM                        & .613                        &.617                             &                             \\
Macro + activity LSTM + static GCN &.616                         &.623                           &.651                            \\    
Temporal GCN-LSTM          & .627               & .629          &           \\
Macro + activity LSTM + Temporal GCN-LSTM~~~  & \textbf{.629}   &\textbf{.633}\\\midrule
\end{tabular}
}
\vspace{-10pt}
\end{table}

\subsection{Performance Study on Classification of User Engagement Trends}

To compare the performance of graph features and graph embedding we consider macroscopic feature prediction as a weaker baseline and LSTM based activity sequence prediction as a stronger baseline, SVM as our baseline classifier and Softmax classifier to showcase the final performance of the classification task. Experiment results are recorded in Table \ref{tab:activity prediction}. 

\smallskip
\noindent \textbf{1. Feature-based model vs. end-to-end neural model.}
Feature-based model can be trained with basic macroscopic features, graph features as well as different embedding features in a 2-step manner. End-to-end model on the other hand, jointly train all model parameters together. The end-to-end model in general learns a better fit for prediction, also more desirable and less complicated.

\smallskip
\noindent \textbf{2. Modeling static vs. temporal graphs.}
Static graphs are aggregated over sessions in the whole 2 weeks observation period, whereas temporal graphs are aggregated per day for each user. While static graphs do not consider evolution of user behavior patterns, temporal graphs model action graphs with time dependency. As a result, we can see in Table \ref{tab:activity prediction} that when modeling the time variant of action graphs prediction performance improves. Meaning that we can successfully capture user behavior patterns with daily evolving action graphs. 

\smallskip
\noindent \textbf{3. End-to-End integration of different models.}
In our 3-way classification task for engagement trend prediction, we experiment different combinations of features. We use LSTM activity sequence embedding as a baseline from previous work \cite{yang2018know}. Our temporal graph model alone is able to outperform LSTM activity sequence model. Adding static graph embedding on top of macroscopic features and activity embedding is able to increase performance. Combining macroscopic features, activity embedding and temporal graph embedding reaches best performance.

\smallskip
\noindent \textbf{4. Ablation study on graph features.}
We also conduct an ablation study of how our explainable graph features improve performance as shown in Table \ref{tab:graph ablation}. Using macroscopic features as baseline, each graph feature adds on different effects and improvements of performance to baseline macroscopic features. A combination of all graph features show that graph features adds  predictive strength to user engagement and also serves as an explanation to why a deeper graph network would work.

\begin{table}[t]
\caption{\textbf{Ablation study on different graph features for engagement trend classification.} We constructed a base model using only macroscopic features and integrate it with different graph features. 
}
\vspace{-0.2cm}
\label{tab:graph ablation}
\scalebox{0.95}{
\begin{tabular}{@{}lc@{}}
\toprule
\textbf{Graph features}                               & \multicolumn{1}{l}{\textbf{F1 score}} \\ \midrule
Macroscopic features                   & .452                                                  \\
Macro + First actions                & .473                                                  \\
Macro + 3-hop paths                     & .484                                                  \\
Macro + End-to-End paths                 & .482                                                  \\
Macro + Strongest elementary cycles    & .482                                                  \\
Macro + All graph features & \textbf{.502}                                                  \\ \bottomrule
\end{tabular}
}
\vspace{-10pt}
\end{table}

\subsection{Results on Active Rate Prediction}
Active rate prediction is a more fine-grained task than engagement level prediction, as it is a regression problem of 14 day range. The same combination of features in engagement trend prediction are experimented. 

\smallskip
\noindent \textbf{1. Temporal graph versus Activity sequence.}
In a more fine-grained task, we can see from Table \ref{tab:active rate pred} that modeling temporal graphs is able to provide a far superior prediction result than baseline activity sequence. The result that such a huge performance jump can be achieved by modeling temporal graphs shows its ability to predict future engagement rate.

\smallskip
\noindent \textbf{2. Feature integration.}
We also integrate multiple feature combinations into our model. A combination of both baseline activity sequence and temporal action graph along with macroscopic features reaches best performance. With information of user engagement that action graph captures, business decisions can be made such as user retention and targeted incentives.

\begin{table}[th]
\caption{\textbf{Performance comparison on active rate prediction.} We report the root mean square error (RMSE) of active rate prediction (regression). Smaller RMSE indicates a better prediction performance on the test set.
}
\vspace{-0.2cm}
\label{tab:active rate pred}
\scalebox{0.95}{
\begin{tabular}{@{}llc@{}}
\toprule
\textbf{Features}                                    & \textbf{RMSE} \\ \midrule
Macro features                                       &  5.30    \\
Graph features                                       &  4.70    \\
Macro + graph features                               &  4.65    \\
Static GCN                                  &  4.25    \\
Activity LSTM                               &  3.69    \\
Macro + activity LSTM                        &  3.32    \\
Macro + activity LSTM + static GCN & 3.23      \\
Temporal GCN                             &2.95      \\ 
Macro + activity LSTM + temporal GCN~~~~~ &\textbf{2.75} \\ \bottomrule
\end{tabular}
}
\vspace{-10pt}
\end{table}

\subsection{Case Study on Graph Features}
\label{sec:feature selection}
Amid many explainable graph features, we are eager to find features that most differentiates levels of user engagement. A simple L1 penalty term can be added to linear kervel SVM classifier to produce sparse coefficients and naturally acts as a feature selection tool. Therefore, variances of features that contributes most to prediction can be extracted after training all graph features with linear SVM. The most predictive graph features of higher order includes k-hop paths, end-to-end paths and cycle strength. Most top 3-hop paths from session start contains chat activities. It is evident that chat activities can imply communication with friends, which subsequently acts as a strong indicator of engagement increase. Top cycles on the other hand are mostly complex structures not simple cycles. This indicates that a well rounded user that is involved with more core functions on Snapchat is more engaged. Similarly, selected end-to-end paths are longer paths, not short simple paths. It implies that user spending more time, uses more functions in each session serves as a strong measure to the users' engagement. As we move on to deeper neural models, the interpretability of simple graph features are the premise to the reason why it works.

%% file: 6-related-work.tex
\section{Related Work}
\label{sec::related_work}

\noindent \textbf{User behavior modeling and prediction.}
Many papers models user behavior and aim to predict if a user will return to the platform in a variety of ways. Return rate prediction is one of the most prevalent and significant task on social media platforms as companies are eagerly trying to find out the secret to their success. A few studies aim to predict churn rate, i.e cease activity on the platform, or user retention\cite{yang2018know, au2003novel, kawale2009churn}. Others come in from a different angle to predict when a user will return \cite{kapoor2014hazard}, if a user will return \cite{lin2018ll} or determine the lifespan of a new user \cite{yang2010activity}.
Another purpose of user behavior modeling leads to the discovery and prediction of user intention. Studies were done on predicting user intent and subsequent behavior, as well as prediction of a user’s intention \cite{cheng2017predicting} and the users purchasing intent \cite{lo2016understanding} on pinterest. Interesting to mention, some studies in the domain focus and models the consumption of a user \cite{benson2016modeling, trouleau2016just}.
Many different methods were used in the task of prediction. Some uses simple logistic regression or gradient boosted tree methods with macroscopic features and reaches good result\cite{lin2018ll, benson2016modeling, althoff2015donor}. Others incorporate time information and utilizes Cox proportional hazard models \cite{kapoor2014hazard, yang2010activity} or Long short-term memory structure \cite{yang2018know} for prediction.

\smallskip
\noindent \textbf{Action sequence modeling.}
There are plenty of publications modeling activity sequence as a graph in the field of search platform and social platforms. However, different from our work that models actions into activity graphs, related literature models Queries and Clickstreams into activity graphs(markov chain graphs). In literatures on search platforms, Queries can be modeled into a sequential graph to predict search success \cite{hassan2010beyond} or produce recommendation \cite{boldi2008query}. Queries graphs are also used to model user search behavior \cite{baeza2005modeling}. Clickstreams on the other hand were also used to model user behavior as a sequential graph/markov chain model on social and web platforms \cite{sadagopan2008characterizing, benevenuto2009characterizing, gunduz2003web, wang2017clickstream}. There are studies done on activity sequence graphs with search queries and clickstream but we have yet to see activity sequence graphs modeled with user activities on a social media platform.

%% file: 7-conclusion.tex
\section{Conclusion}

In this paper, we explore and analyze user behavior of Snapchat's new user through action graphs. Following analysis insights we developed a robust and predictive action graph modeling framework. In our analysis we provide valuable insights to understand new user patterns after initial registration and likelihood of future engagement. While our models and analysis focuses on Snapchat data, similar techniques and models can be applied to other real world social media or online platforms.
Our action graph prediction framework is \textit{deployed in Snap Inc.} to deliver analysis results and benefits according business and production decisions which includes but not limited to user modeling, growth and retention. Possible future work can be done on connecting users ego-network with its behavioral pattern.
